\documentclass[10pt,conference]{IEEEtran}

\usepackage{cite}
\ifCLASSINFOpdf
   \usepackage[pdftex]{graphicx}
   \graphicspath{{/}}
   \DeclareGraphicsExtensions{.pdf,.jpeg,.png}
\else
   \usepackage[dvips]{graphicx}
   \graphicspath{{../}}
   \DeclareGraphicsExtensions{.eps}
\fi
\usepackage[cmex10]{amsmath}
\usepackage{amsthm}

\usepackage{algorithmic}
\usepackage{array}
\ifCLASSOPTIONcompsoc
  \usepackage[caption=false,font=normalsize,labelfont=sf,textfont=sf]{subfig}
\else
  \usepackage[caption=false,font=footnotesize]{subfig}
\fi
\usepackage{url}

\begin{document}
%
\title{A Structural Graph-Based Method for MRI Analysis}

\newif\iffinal
\finaltrue
\newcommand{\jemsid}{10}


\iffinal

\author{\IEEEauthorblockN{Larissa de O. Penteado, Mateus Riva\\and Roberto M. Cesar Jr.}
\IEEEauthorblockA{Institute of Mathematics and Statistic \\
University of S\~ao Paulo\\
Emails: \{lariop, mriva, cesar\}@ime.usp.br}
\and
\IEEEauthorblockN{Marcelo Straus Takahashi, Lisa Suzuki\\and
Carlos A. Moreira-Filho}
\IEEEauthorblockA{Department of Pediatrics and Instituto da Crian\c{c}a -- HC-FM\\
University of S\~ao Paulo\\
Emails: \{marcelo.straus, lisa.suzuki, carlos.moreira\}@hc.fm.usp.br}}


\else
  \author{SIBGRAPI paper ID: \jemsid \\ }
\fi

\maketitle

\begin{abstract}
The importance of imaging exams, such as Magnetic Resonance Imaging (MRI), for the diagnostic and follow-up of pediatric pathologies and the assessment of anatomical structures' development has been increasingly highlighted in recent times. Manual analysis of MRIs is time-consuming, subjective, and requires significant expertise. To mitigate this, automatic techniques are necessary. Most techniques focus on adult subjects, while pediatric MRI has specific challenges such as the ongoing anatomical and histological changes related to normal development of the organs, reduced signal-to-noise ratio due to the smaller bodies, motion artifacts and cooperation issues, especially in long exams, which can in many cases preclude common analysis methods developed for use in adults. Therefore, the development of a robust technique to aid in pediatric MRI analysis is necessary. This paper presents the current development of a new method based on the learning and matching of structural relational graphs (SRGs). The experiments were performed on liver MRI sequences of one patient from ICr-HC-FMUSP, and preliminary results showcased the viability of the project. Future experiments are expected to culminate with an application for pediatric liver substructure and brain tumor segmentation.
\end{abstract}


\IEEEpeerreviewmaketitle

\section{Introduction}
\label{sec:introduction}
Magnetic Resonance Imaging (MRI) is an important tool for examination and diagnosis of a wide array of pathologies under diverse circumstances. Its analysis has several applications, such as the study of Alzheimer's disease evolution~\cite{Wang16-JAD}, multiple sclerosis lesions~\cite{Garcia13-MIA}, and bipolar disorder~\cite{Hibar18-MP}, to name a few examples.

However, manual analysis is often subjective~\cite{Morel16-NR} and time-intensive. Thus, fast and automatic methods of MRI analysis are required. As several studies are based on the segmentation of different tissues or structures in the body, and the amount of patient data available grows significantly, automatic or semi-automatic segmentation of regions of interest becomes essential.

While several methods for the analysis of adult MRI have been developed over the years, relatively few have been developed for pediatric MRI~\cite{Devi15-CBM}. Pediatric images present particular challenges~\cite{Devi15-CBM,Neubauer13-PR}, such as the inversion of the contrasts in white and gray matter, patient motion artifacts~\cite{Devi15-CBM} and significant corporal variation among individuals with small age differences~\cite{Huppi98-AN}.

Additionally, most segmentation methods focus on healthy patients, or on a single anomaly (such as a specific type of tumor)~\cite{harish2017-Extensive}. There is a demand for methods that are capable of dealing with a wide range of abnormalities -- extraneous elements such as tumors or missing elements such as organs removed by surgery -- in the corporal structure without significant retraining or adjustment.

When dealing with structured scenes, that is, scenes where the objects of interest obey some sort of predictable spatial structure (such as the interior of the body), the usage of structural information to aid segmentation and analysis has been shown to allow for significant increases in accuracy~\cite{Colliot06-PR,Fouquier12-CVIU,Morel16-EMBC,Nempont13-IS}.


The goal of this project is to develop a structural (graph-based) method for the analysis of pediatric images. This method will be applied to real MRI data in order to solve active problems in the area.

\section{Methods and Approach}
\label{sec:methods}
\begin{figure}[h!]
\centering
\includegraphics[width = 1\linewidth]{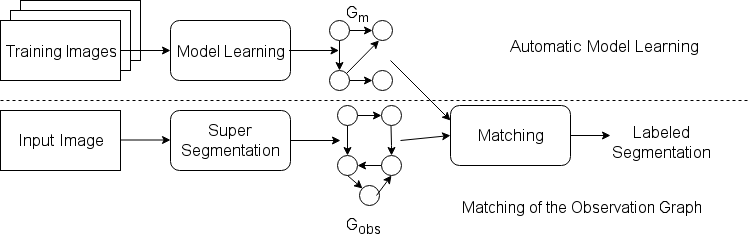}
\caption{Pipeline showing the steps of the segmentation task.}
\label{fig:pipe}
\end{figure}

The project will involve the design of a suitable graph-based structural model for image segmentation and body parts recognition. In order to accurately model and represent the knowledge found in MRI data, proper understanding of the structural components of the images is required. As shown by \cite{Colliot06-PR,Fouquier12-CVIU,Morel16-EMBC,Nempont13-IS}, structural knowledge is of primary importance for the interpretation of images where intensity and shape may not be sufficient data for an adequate analysis, such as abdominal and brain MRIs. To mitigate this deficiency, we propose the usage of graphs (and, potentially, hypergraphs) for the representation of anatomical structures, where each vertex represents a region or structure of interest and the edges represent relations between these (such as relative spatial positioning). Our challenge will be the application of this knowledge in a way that guides and enhances the image interpretation process.  Figure \ref{fig:pipe} presents the proposed pipeline to be followed in order to segment and recognize structures of interest in the MRI sequence.

\subsection{The Statistical-Relational Graph}
\label{sec:srg}
The Statistical-Relational Graph (SRG)~\cite{Graciano12-THESIS} is an hybrid structured scene representation technique based on Attributed Relational Graphs, where vertices of a graph represent primitive elements of the scene and edges represent their structural relationship. Both vertices and edges contain attributes which describe the element or relationship they represent.

For recognition of structured scenes, we first build a graph which represents the template for a given scene, with a set of primitive and relational attributes, called the \emph{model graph}. Object recognition is thus a matter of building an equivalent or comparable graph from the observed input, the \emph{observation graph}, and attempting to match both graphs. 

Structural-relational graphs have been successfully applied to object tracking in digital image sequences~\cite{Graciano07-SIBGRAPI,Morimitsu15-THESIS}, automatic image segmentation~\cite{Graciano12-THESIS} and interactive image segmentation~\cite{Noma08-ARXIV, Noma12-PR}.

In their thesis, Graciano~\cite{Graciano12-THESIS} demonstrates this approach's robustness in the segmentation of abdominal organs in normal, adult patient MRIs, utilizing classic probabilistic pattern recognition techniques to describe organ tissues while simultaneously considering the spatial relationship between distinct organs in order to improve segmentation quality. We believe this method may be further developed for the automatic segmentation and recognition of regions of interest in pediatric MRIs.

Primitive elements, in the context of MRI analysis, may be organs, veins, tumors, or similar regions of interest such as hepatic divisions. Discriminative attributes for these elements may include mean intensity, total volume, and position in the body. Structural relationships describes the connections between distinct primitive elements, such as relative position and distance, relative intensity and relative size.

\begin{figure}
\centering
\includegraphics[width=1.0\linewidth]{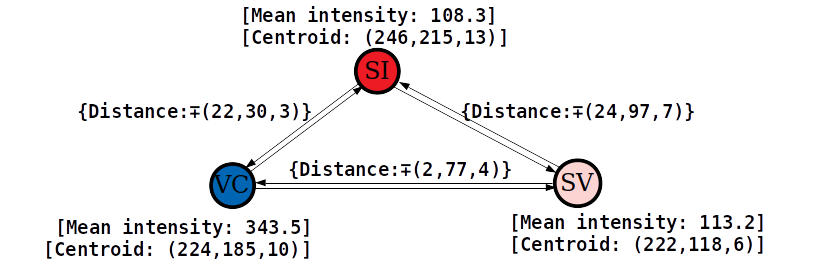}
\caption{Schema of a simple, example SRG. Vertex attributes are between square brackets. Edge attributes are between curly braces.}
\label{fig:srg_example}
\end{figure}

Figure \ref{fig:srg_example} presents an example of a SRG. The vertices represent some abdominal structures of interest: Segment I (SI), Segment V (SV) and the Vena Cava (VC). The vertex attributes are the `centroid' and the `mean intensity', calculated from the MRI for each of the structures (vertices). The edge attribute is `distance', which represents the vector between the centroid of the two structures in the edge's vertices.

\subsection{Model Description}
\label{sec:model_description}

In order to perform our tasks of segmentation and recognition, we must first build a model capable of extracting and properly representing the structures and objects of interest. This model must be able to discriminate precisely between structures, using both statistic attributes from the structures and features derives from the relationships between them.

\subsubsection{Conceptual modeling}
\label{ssec:conceptual}

In order to properly segment and recognize the structures of interest, the following attributes of each structure must be considered: \textbf{i)} Image intensity values; \textbf{ii)} volume; and \textbf{iii)} absolute position in the body.

Additionally, the relationships between a structure and all the others must also be taken into consideration, with the following attributes being most relevant: \textbf{i)} relative distance, position and orientation; \textbf{ii)} differences in volume; and \textbf{iii)} contrast in intensity values.

We believe the combination of these statistical and relational attributes represent an accurate approximation of the discriminative factors required for proper segmentation and recognition of the relevant structures. Previous works have shown successful results produced by such attributes~\cite{Graciano12-THESIS, Colliot06-PR,Graciano14-ICPR,Morimitsu16-CVIU}.

\subsubsection{Computational modeling}
\label{ssec:computational_model}

Considering the conceptual model proposed in Section~\ref{ssec:conceptual}, we then define a single data structure, known as Statistical-Relational Graph (SRG), which will be used to represent the conceptual model.

Formally, the SRG is defined as a quadruple $G = (V, E, A_V, A_E)$ where:
\begin{itemize}
\item $V$ is the set of vertices in $G$. Each vertex represents a single part of the object of interest;
\item $E$ is the set of edges in $G$. Each edge represents a relationship between a pair of parts of the object;
\item $A_V$ is the set of vertex attributes and their respective probability distributions. Each element in $A_V$ maps to a single vertex in $V$ and contains that vertex's attributes;
\item $A_E$ is the set of edge attributes and their respective probability distributions. Each element in $A_E$ maps to a single edge in $E$ and contains that edge's attributes.
\end{itemize}

Three vertex attributes are computed for each vertex. Considering the structure they represent, these attributes are \textbf{i)} the average of the intensity values in the structure; \textbf{ii)} the total volume of the structure; and \textbf{iii)} the position of the centroid of the structure.

The volume attribute, in particular, may hinder initial matching efforts due to the difference between the volumes of the model vertices and the observation vertices, as the latter are much less voluminous than the former. However, this project aims to explore a way to utilize volume and, potentially, orientation information to aid the matching of a SRG.

Three edge attributes are computed for each edge. Considering the structures represented by the connected edges, these attributes are \textbf{i)} the vector between the centroids of each structure; \textbf{ii)} the proportional difference in volume between the structures; and \textbf{iii)} the contrast (that is, difference) between the average intensities of each structure.

\subsubsection{Graph creation}
\label{sssec:graph_creation}
From an annotated volume, an SRG is created as follows: for each distinct label, a vertex is created. The corresponding attributes are then computed for each label. 
Once all vertices are created, the graph is then fully connected, and edge attributes are computed.

A model graph is created from manually annotated images. One of this project's goals is to find the best possible model graph from a set of such images. The sets of vertices and edges ($(V, E)$) will be defined \textit{a priori}. For example, $V$ will be provided from an anatomical atlas and $E$ will have a specified topology (e.g. fully connected). Then, learning algorithms will be applied in order to learn  the attributes' distributions in the statistical model. Initially, we'll consider a few multivariate distributions, such as Gaussian, for both parameters.

A super-observation graph is created from an automatic super-segmentation algorithm such as watershed. From the super-observation graph, regions may be joined to form an observation graph of the same size as the model graph.

\subsubsection{Graph matching}
\label{sssec:graph_matching}
In order to compare two SRGs, a matching solution must be generated and evaluated. A solution $S$, which matches a super observation graph $G_{super}$ with $|V_{super}| = n_{super}$ vertices to a model graph $G_M$ with $|V_M| = n$ vertices, is defined as
\begin{equation}
	S = [s_1, s_2, ..., s_{n_{super}}], s_i \in \{1,n\}
\end{equation}

From $S$, all regions with the same prediction are joined and the observation graph $G_{obs}$, with $|V_{obs}| = n$, is created. To evaluate the quality of $S$, a cost function must be established. The cost of a solution $S$ is defined as
\begin{equation}
	C(S) = \alpha \frac{1}{n} \sum_{j=1}^{n}c_V(s_j) + (1-\alpha)\frac{1}{{n}^2} \sum_{j=1}^{n}\sum_{k=1}^{n}c_E(s_j,s_k)
\end{equation}
where $\alpha \in [0,1]$ is a the weight of the verticial cost; the costs between vertices $j$ in $G_{obs}$ and $s_j$ in $G_M$, $c_V(s_j)$, is defined as
\begin{equation}
	c_V(s_j) = \sum_{a\in A_V}{\alpha_a d_a({s_j}_a)}
\end{equation}
and the cost between the edges $\{j,k\}$ in $G_{obs}$ and $\{s_j,s_k\}$ in $G_M$, $c_E(s_j,s_k)$, is defined as
\begin{equation}
	c_E(s_j,s_k) = \sum_{a\in A_E}{\alpha_a d_a(\{s_j\}_a, \{s_k\}_a)}
\end{equation}
with $d_a$ being the distance function corresponding to each attribute $a$ and $\alpha_a$ being a customizable parameter, used for tuning the weight of the distance between each attribute $a$ in the final cost. Note that $\sum_{a\in A_E}{\alpha_a} = 1$.

Given a super-observation SRG $G_{super}$, an initial solution may be built using a simple greedy algorithm,
where each super-observation vertex $j \in V_{super}$ is matched to the model vertex $i \in V_M$ with the lowest cost, that is, $s_j = argmin_{i=0}^{n}(c_V(j,i)), j=0,1,...,n_{super}$.

\section{Experiments}
\label{sec:experiments}
\subsection{Data Acquisition}
The main source of data for this project is the ICr-HC-FMUSP. All acquisitions were performed using ICr's Philips Ingenia (1.5T) machine. Abdominal T2-weighted MRI of nine patients, ranging from four to eighteen years old, were used for experiments in liver segmentation; among these patients, six were affected with hepatic iron buildup and three were healthy at the time of acquisition. From the same source, there is also brain T1, T2 and Diffusion MRI of three healthy patients. Their ages were 7, 8 months and 13 years at the time of acquisition. Further patients will be introduced to the dataset, with ages ranging from 6 months old to 13 years old, which may improve the method's efficiency.

\subsection{Exploratory SRG Experiments}
Exploratory experiments were conducted utilizing only \emph{mean intensity}, \emph{centroid} and \emph{distance} features and data from a single patient from the abdominal MRI dataset. The goal of these experiments was to determine the usefulness of the method for solving the problem in the given data, and spot and solve any potential problems that may arise in a more complete implementation.

The model graph was created from the annotated patient's data. The observation graph was built from an watershed segmentation of the morphological gradient of the full volume
; the regions produced by this segmentation were used as the vertices of the observation graph. We then matched both graphs using the simple greedy algorithm described in Section~\ref{sssec:graph_creation}. We explored several parameters for the $\alpha_a$ tuning parameters.

\subsection{Results and Discussion}
\begin{figure}
	\centering
	\begin{tabular}{cc}
		\includegraphics[width=0.45\linewidth]{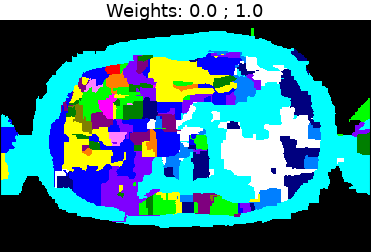}&\includegraphics[width=0.45\linewidth]{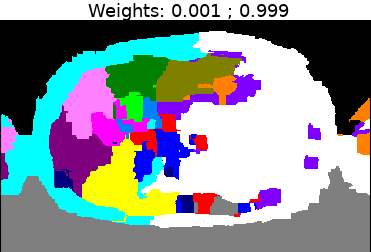}\\
        (a)&(b)\\
		\includegraphics[width=0.45\linewidth]{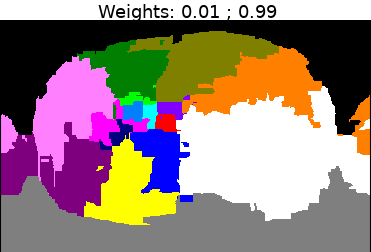}&\includegraphics[width=0.45\linewidth]{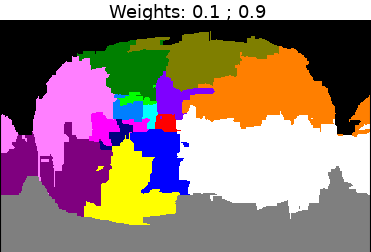}\\
        (c)&(d)\\
		\includegraphics[width=0.45\linewidth]{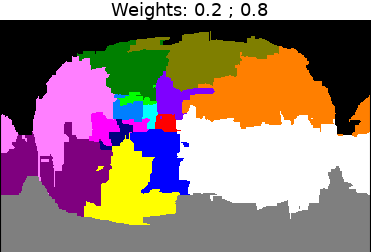}&\includegraphics[width=0.45\linewidth]{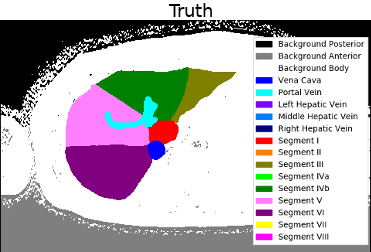}\\
        (e)&(f)
	\end{tabular}
    \caption{Selected results from the exploratory experiments on SRG. All images are from the same slice. Weights given are for the \emph{centroid} and the \emph{mean intensity}, respectively. All results for \emph{centroid} weights above 0.2 are similar.}
    \label{fig:preliminary_results}
\end{figure}

The results of the exploration of different vertex attribute weights, and their effects both on establishing an initial, greedy solution and computing the total cost of the solution, are shown on Table~\ref{tab:preliminary_results}. Selected results are displayed in Figure~\ref{fig:preliminary_results}.

\begin{table}
	\centering
	\caption{Results of the exploratory experiments on SRG.}
	\label{tab:preliminary_results}
    \begin{tabular}{c|c|c}
    	\textbf{Centroid $\alpha$} & \textbf{Intensity $\alpha$} & \textbf{Cost}\\
        \hline
        0 & 1 & $39.659.347$ \\
        0.001 & 0.999 & $29.912.702$ \\
        0.005 & 0.995 & $23.693.941$ \\
        0.01 & 0.99 & $23.064.714$ \\
        0.02 & 0.98 & $22.803.427$ \\
        0.1 & 0.9 & $22.720.578$ \\
        0.2 & 0.8 & $22.718.200$ \\
        0.5 & 0.5 & $22.718.200$ \\
        1 & 0 & $22.718.200$ \\
    \end{tabular}
\end{table}

We can notice that the \emph{centroid} attribute dominates the \emph{mean intensity}; indeed, for all weights of the \emph{intensity} attribute below $0.8$, including $0.0$, there are no differences in cost. While this may point to a significant advantage in using the \emph{centroid}, this may also be the result of a deficiency of more discriminant attributes (such as volume), of improper segmentation in the watershed step, and of improper normalization of the attributes. These results have paved the way for the next set of experiments, which will explore different sets of segmentation algorithms, region joining in the observation graph, and attribute normalization techniques.

\section{Conclusion}
\label{sec:conclusion}
We proposed a method for automatic segmentation and recognition of structures of interest in pediatric MRI. As stated before, this is an active and interesting problem with very clear real-world applications. The success of this project will both help pave the way for further applications using SRGs in medical imaging, and provide the scientific and medical community with a useful product, capable of aiding the radiologist in diagnostic and analysis of liver MRIs.

The proposed framework is already under implementation, and relevant data has been acquired and prepared for use. Previous work on similar methods show that this approach has potential for fast and accurate segmentation and recognition of structures, as well as high extensibility and capability for improvement and modification, such as transposing the proposed method to a different domain.

However, as the preliminary results of the exploratory experiments show, there are significant avenues for improvement that must be explored. Improving the quality of the observation graph, through the use of more discriminant attributes, better algorithms for building solutions, and improved super-segmentation approaches, is of utmost importance.

\section*{Acknowledgment}
The authors would like to thank the Funda\c{c}\~ao de Apoio \`a Pesquisa do Estado de S\~ao Paulo (FAPESP), grants \#2017/09465-7, \#2018/07386-5, \#2017/50236-1 and \#2015/22308-2; and also the Coordena\c{c}\~ao de Aperfei\c{c}oamento de Pessoal de N\'ivel Superior (CAPES), grant \#1767508.

\bibliographystyle{IEEEtran}
\bibliography{refs}

\begin{thebibliography}{10}
\providecommand{\url}[1]{#1}
\csname url@samestyle\endcsname
\providecommand{\newblock}{\relax}
\providecommand{\bibinfo}[2]{#2}
\providecommand{\BIBentrySTDinterwordspacing}{\spaceskip=0pt\relax}
\providecommand{\BIBentryALTinterwordstretchfactor}{4}
\providecommand{\BIBentryALTinterwordspacing}{\spaceskip=\fontdimen2\font plus
\BIBentryALTinterwordstretchfactor\fontdimen3\font minus
  \fontdimen4\font\relax}
\providecommand{\BIBforeignlanguage}[2]{{%
\expandafter\ifx\csname l@#1\endcsname\relax
\typeout{** WARNING: IEEEtran.bst: No hyphenation pattern has been}%
\typeout{** loaded for the language `#1'. Using the pattern for}%
\typeout{** the default language instead.}%
\else
\language=\csname l@#1\endcsname
\fi
#2}}
\providecommand{\BIBdecl}{\relax}
\BIBdecl

\bibitem{Wang16-JAD}
S.~Wang, Y.~Zhang, G.~Liu, P.~Phillips, and T.-F. Yuan, ``Detection of
  alzheimer’s disease by three-dimensional displacement field estimation in
  structural magnetic resonance imaging,'' \emph{Journal of Alzheimer's
  Disease}, vol.~50, no.~1, pp. 233--248, 2016.

\bibitem{Garcia13-MIA}
D.~Garc{\'\i}a-Lorenzo, S.~Francis, S.~Narayanan, D.~L. Arnold, and D.~L.
  Collins, ``Review of automatic segmentation methods of multiple sclerosis
  white matter lesions on conventional magnetic resonance imaging,''
  \emph{Medical Image Analysis}, vol.~17, no.~1, pp. 1--18, 2013.

\bibitem{Hibar18-MP}
D.~Hibar, L.~Westlye, N.~Doan, N.~Jahanshad, J.~Cheung, C.~Ching, A.~Versace,
  A.~Bilderbeck, A.~Uhlmann, B.~Mwangi \emph{et~al.}, ``Cortical abnormalities
  in bipolar disorder: an mri analysis of 6503 individuals from the enigma
  bipolar disorder working group,'' \emph{Molecular psychiatry}, vol.~23,
  no.~4, p. 932, 2018.

\bibitem{Morel16-NR}
B.~{Morel}, G.~{Antoni}, J.~{Teglas}, I.~{Bloch}, and C.~{Adamsbaum},
  ``Neonatal brain {MRI}: How reliable is the radiologist's eye?''
  \emph{Neuroradiology}, vol.~58, pp. 189--193, 2016.

\bibitem{Devi15-CBM}
C.~N. Devi, A.~Chandrasekharan, V.~Sundararaman, and Z.~C. Alex, ``Neonatal
  brain mri segmentation: A review,'' \emph{Computers in biology and medicine},
  vol.~64, pp. 163--178, 2015.

\bibitem{Neubauer13-PR}
H.~Neubauer, T.~Pabst, A.~Dick, W.~Machann, L.~Evangelista, C.~Wirth,
  H.~K{\"o}stler, D.~Hahn, and M.~Beer, ``Small-bowel mri in children and young
  adults with crohn disease: retrospective head-to-head comparison of
  contrast-enhanced and diffusion-weighted mri,'' \emph{Pediatric radiology},
  vol.~43, no.~1, pp. 103--114, 2013.

\bibitem{Huppi98-AN}
P.~S. H{\"u}ppi, S.~Warfield, R.~Kikinis, P.~D. Barnes, G.~P. Zientara, F.~A.
  Jolesz, M.~K. Tsuji, and J.~J. Volpe, ``Quantitative magnetic resonance
  imaging of brain development in premature and mature newborns,'' \emph{Annals
  of neurology}, vol.~43, no.~2, pp. 224--235, 1998.

\bibitem{harish2017-Extensive}
S.~Harish, G.~A. Ahammed, and R.~Banu, ``An extensive research survey on brain
  mri enhancement, segmentation and classification,'' in \emph{Electrical,
  Electronics, Communication, Computer, and Optimization Techniques (ICEECCOT),
  2017 International Conference on}.\hskip 1em plus 0.5em minus 0.4em\relax
  IEEE, 2017, pp. 1--8.

\bibitem{Colliot06-PR}
O.~Colliot, O.~Camara, and I.~Bloch, ``Integration of fuzzy spatial relations
  in deformable models—application to brain mri segmentation,'' \emph{Pattern
  Recognition}, vol.~39, no.~8, pp. 1401--1414, 2006.

\bibitem{Fouquier12-CVIU}
G.~Fouquier, J.~Atif, and I.~Bloch, ``Sequential model-based segmentation and
  recognition of image structures driven by visual features and spatial
  relations,'' \emph{Computer Vision and Image Understanding}, vol. 116, no.~1,
  pp. 146--165, 2012.

\bibitem{Morel16-EMBC}
B.~Morel, Y.~Xu, A.~Virzi, T.~G{\'e}raud, C.~Adamsbaum, and I.~Bloch, ``A
  challenging issue: Detection of white matter hyperintensities in neonatal
  brain mri,'' in \emph{Engineering in Medicine and Biology Society (EMBC),
  2016 IEEE 38th Annual International Conference of the}.\hskip 1em plus 0.5em
  minus 0.4em\relax IEEE, 2016, pp. 93--96.

\bibitem{Nempont13-IS}
O.~Nempont, J.~Atif, and I.~Bloch, ``A constraint propagation approach to
  structural model based image segmentation and recognition,''
  \emph{Information Sciences}, vol. 246, pp. 1--27, 2013.

\bibitem{Graciano12-THESIS}
A.~B.~V. Graciano, ``Modelagem e reconhecimento de objetos estruturados: uma
  abordagem estat{\'\i}stico-estrutural,'' Ph.D. dissertation, University of
  S{\~a}o Paulo, 2012.

\bibitem{Graciano07-SIBGRAPI}
A.~B. Graciano, R.~M. Cesar-Jr, and I.~Bloch, ``Graph-based object tracking
  using structural pattern recognition,'' in \emph{Computer Graphics and Image
  Processing, 2007. SIBGRAPI 2007. XX Brazilian Symposium on}.\hskip 1em plus
  0.5em minus 0.4em\relax IEEE, 2007, pp. 179--186.

\bibitem{Morimitsu15-THESIS}
H.~Morimitsu, ``A graph-based approach for online multi-object tracking in
  structured videos with an application to action recognition,'' Ph.D.
  dissertation, University of S{\~a}o Paulo, 2015.

\bibitem{Noma08-ARXIV}
A.~Noma, A.~B. Graciano, L.~A. Consularo, R.~M. Cesar-Jr, and I.~Bloch, ``A new
  algorithm for interactive structural image segmentation,'' \emph{arXiv
  preprint arXiv:0805.1854}, 2008.

\bibitem{Noma12-PR}
A.~Noma, A.~B. Graciano, R.~M. Cesar~Jr, L.~A. Consularo, and I.~Bloch,
  ``Interactive image segmentation by matching attributed relational graphs,''
  \emph{Pattern Recognition}, vol.~45, no.~3, pp. 1159--1179, 2012.

\bibitem{Graciano14-ICPR}
A.~B.~V. {Graciano}, R.~M. {Cesar Jr.}, and I.~{Bloch}, ``Modeling and
  recognition of structured objects: a statistical-relational approach,'' in
  \emph{Stockholm. Proc. Workshop FEAST - ICPR 2014}.\hskip 1em plus 0.5em
  minus 0.4em\relax IEEE, 2014.

\bibitem{Morimitsu16-CVIU}
H.~Morimitsu, I.~Bloch, and R.~M. Cesar-Jr, ``Exploring structure for long-term
  tracking of multiple objects in sports videos,'' \emph{Computer Vision and
  Image Understanding}, 2016.

\end{thebibliography}

\end{document}